\documentclass[conference]{IEEEtran}
\IEEEoverridecommandlockouts
\usepackage{cite}
\usepackage{amsmath,amssymb,amsfonts}
\usepackage{graphicx}
\usepackage{textcomp}
\usepackage{xcolor}
\usepackage[linesnumbered,ruled,vlined]{algorithm2e}
\usepackage{hyperref}
\usepackage{cleveref}
\usepackage{siunitx}
\usepackage{subcaption}
\usepackage{booktabs}
\usepackage{multirow}
\usepackage{float}
\usepackage{placeins}

\newcommand{\comparisontable}{
    \begin{table}[ht]
    \caption{
    Comparison of low-rank and pruning methods for ResNet-50 (on ImageNet-2012) and VGG16 (on CIFAR10). 'Difference to Baseline' indicates validation top 1 performance relative to the original full-rank model in each study, with positive values denoting better performance. AB training used a global batch size of 4,096 for ImageNet and 1,024 for CIFAR10, achieving maximum top 1 accuracies of 75.67\% and 91.87\% respectively. The estimated communication reduction (ECR) is defined by \Cref{eq:ecr}. AB training's ECR assumes independent groups do not utilize the compute system's interconnect.}
    \label{tab:comparison}
    \centering
    \begin{tabular}{@{}ll|rrr@{}}
    & Method & \begin{tabular}[c]{@{}r@{}}Difference \\ to Baseline\end{tabular} & \begin{tabular}[c]{@{}r@{}}Compression \\ Ratio\end{tabular} & ECR\\ \midrule
    \multirow{9}{*}{\rotatebox[origin=c]{90}{\begin{tabular}[c]{@{}c@{}}ResNet-50 \\ ImageNet-2012 \end{tabular}}} & AB & \textbf{+1.55 \%} & $1.39:1$  & \textbf{73.29 \%} \\
    & AB - No Groups & -0.02 \% & $1.27:1$  & 15.94 \%\\
    & OIALR~\cite{coquelin_harnessing_2024} & -1.72 \% & $1.21:1$  & 63.64 \%\\
    & DLRT~\cite{schotthofer_low-rank_2022} & -0.56 \% & $1.85:1$  & 64.35 \%\\
    & PP-1~\cite{singh_play_2019} & -0.20 \% & $2.26:1$  & 41.81 \%\\
    & CP~\cite{he_channel_2017} & -1.40 \% & $2.00:1$  & 37.5 \%\\
    & SFP~\cite{he_soft_2018} & -0.20 \% & $2.39:1$ & 43.62 \%\\
    & ThiNet~\cite{luo_thinet_2017} & -1.50 \% & \textbf{2.71 : 1}  & 47.32 \%\\
    & ABCPrune~\cite{lin2020abcpruning} & -2.15 \% & 1.84 : 1  & 34.24 \%\\
    \midrule
    \multirow{6}{*}{\rotatebox[origin=c]{90}{\begin{tabular}[c]{@{}c@{}}VGG16 \\ CIFAR-10 \end{tabular}}} &  
    AB & -0.23 \% & \textbf{44.14 : 1}  & 65.60 \%\\
      & AB - No Groups & -0.89 \% & 36.63 : 1  & \textbf{72.95 \%}\\
      & OIALR~\cite{coquelin_harnessing_2024} & \textbf{+0.10 \%} & 3.70 : 1  & 64.59 \%\\
      & DLRT~\cite{schotthofer_low-rank_2022} & -1.89 \% & 1.79 : 1   & 16.88 \%\\
      & ABCPrune~\cite{lin2020abcpruning} & +0.06 \% & 8.83 : 1  & 66.51 \%\\ 
      & ICP~\cite{chang2023icpprune} & -0.31 \% & 27.25 : 1  & 72.25 \%\\ 
     \bottomrule
    \end{tabular}
    \end{table}  
}
\newcommand{\scalingtable}{

\begin{table*}[ht]
\caption{Results from both scaling experiments. Scaled traffic reports the average interconnect traffic across each training run estimated as the number of trainable parameters in \si{\giga\byte} times the number of synchronizations in an epoch divided by the percentage of time taken for a traditional backwards pass and the time taken per epoch. The percentage of time taken for a backwards pass in a traditional DP training step with 4 nodes (16 GPUs) was measured as 53.84\%. The scaled interconnect traffic is limited to the bandwidth available on each node, \SI{25}{\giga\byte / \s}. Compression results show the model compression ratio at the end of training relative to the full-rank model. Time to train shows job wall-clock time. Bold values indicate the most favorable results between AB, AB - No Groups, and traditional DDP training.}
\label{tab:scaling}
\centering
\begin{tabular}{@{}c|r|rrr|rrr@{}}  
\toprule
\multicolumn{8}{c}{Constant Local Batch Size} \\ \midrule
     &  & \multicolumn{3}{c|}{ViT B/16} & \multicolumn{3}{c}{ResNet-50} \\
Batch Size    &  & AB & AB - No Groups & Traditional DDP & AB & AB - No Groups & Traditional DDP \\ \midrule
\multirow{4}{*}{2k}     & Scaled Traffic, \si{\giga\byte / \s} & \textbf{2.81 $\pm$ 0.08} & 5.14 $\pm$ 0.19 & 8.87 $\pm$ 0.00 & \textbf{1.11 $\pm$ 0.00} & 2.87 $\pm$ 0.01 & 3.83 $\pm$ 0.01 \\
    & Compression   & 2.52 : 1 & \textbf{2.83 : 1} & 1 : 1 & \textbf{1.29 : 1} & 1.27 : 1 & 1 : 1 \\
     & Time to Train, \si{\min} & \textbf{435.80 $\pm$ 7.36} & 442.94 $\pm$ 8.52 & 436.38 $\pm$ 0.35 & 368.15 $\pm$ 0.80 & \textbf{364.91 $\pm$ 0.53} & 367.36 $\pm$ 0.36 \\
     & Validation Top 1, \% & 70.24 $\pm$ 0.36 & 68.31 $\pm$ 0.07 & \textbf{70.73 $\pm$ 0.44} & \textbf{75.67 $\pm$ 0.10} & 74.42 $\pm$ 0.04 & 74.43 $\pm$ 0.06 \\ \midrule
\multirow{4}{*}{4k}     & Scaled Traffic, \si{\giga\byte / \s} & \textbf{2.83 $\pm$ 0.06} & 5.72 $\pm$ 0.08 & 8.87 $\pm$ 0.10 & \textbf{1.04 $\pm$ 0.05} & 2.67 $\pm$ 0.04 & 3.85 $\pm$ 0.01 \\
    & Compression   & 2.09 : 1 & \textbf{2.22 : 1} & 1 : 1 & \textbf{1.39 : 1} & 1.27 : 1 & 1 : 1 \\
     & Time to Train, \si{\min} & \textbf{220.85 $\pm$ 0.88} & 224.35 $\pm$ 1.22 & 249.03 $\pm$ 31.01 & 200.69 $\pm$ 10.30 & 195.46 $\pm$ 1.30 & \textbf{195.19 $\pm$ 0.59} \\ 
     & Validation Top 1, \% & \textbf{70.90 $\pm$ 0.44} & 67.81 $\pm$ 0.49 & 69.15 $\pm$ 0.00 & \textbf{74.61 $\pm$ 0.06} & 73.68 $\pm$ 0.12 & 73.67 $\pm$ 0.05 \\ \midrule
\multirow{4}{*}{8k}     & Scaled Traffic, \si{\giga\byte / \s} & \textbf{2.65 $\pm$ 0.04} & 5.93 $\pm$ 0.13 & 8.51 $\pm$ 0.00 & \textbf{0.91 $\pm$ 0.03} & 2.71 $\pm$ 0.02 & 3.71 $\pm$ 0.09 \\
    & Compression   & \textbf{2.02 : 1} & 1.98 : 1 & 1 : 1 & \textbf{1.31 : 1} & 1.16 : 1 & 1 : 1 \\
     & Time to Train, \si{\min}  & \textbf{115.65 $\pm$ 1.37} & 121.95 $\pm$ 1.95 & 121.78 $\pm$ 1.02 & 108.19 $\pm$ 4.94 & \textbf{104.70 $\pm$ 0.52} & 105.77 $\pm$ 1.69 \\ 
     & Validation Top 1, \% & \textbf{68.92 $\pm$ 0.10} & 65.70 $\pm$ 0.14 & 67.20 $\pm$ 0.07 & \textbf{73.66 $\pm$ 0.03} & 73.04 $\pm$ 0.15 & 73.02 $\pm$ 0.10 \\ \midrule
\multirow{4}{*}{16k}     & Scaled Traffic, \si{\giga\byte / \s} & \textbf{2.54 $\pm$ 0.03} & 4.91 $\pm$ 0.67 & 8.25 $\pm$ 0.18 & \textbf{0.87 $\pm$ 0.01} & 2.56 $\pm$ 0.04 & 3.63 $\pm$ 0.23 \\
   & Compression   & \textbf{1.86 : 1} & \textbf{1.86 : 1} & 1 : 1 & \textbf{1.19 : 1} & 1.10 : 1 & 1 : 1 \\
     & Time to Train, \si{\min}  & 64.83 $\pm$ 1.96 & 79.15 $\pm$ 9.31 & \textbf{63.53 $\pm$ 1.10} & \textbf{58.61 $\pm$ 0.07} & 59.54 $\pm$ 0.68 & 59.27 $\pm$ 0.10 \\ 
     & Validation Top 1, \% & \textbf{64.20 $\pm$ 0.01} & 61.17 $\pm$ 0.33 & 63.68 $\pm$ 0.09 & \textbf{72.66 $\pm$ 0.04} & 72.60 $\pm$ 0.10 & 72.32 $\pm$ 0.03 \\ \midrule
\multirow{4}{*}{32k}     & Scaled Traffic, \si{\giga\byte / \s} & \textbf{2.27 $\pm$ 0.41} & 4.63 $\pm$ 0.51 & 7.52 $\pm$ 0.36 & \textbf{0.78 $\pm$ 0.01} & 2.17 $\pm$ 0.04 & 3.00 $\pm$ 0.06  \\
   & Compression   & 1.80 : 1 & \textbf{1.84 : 1} & 1 : 1 & \textbf{1.20 : 1} & 1.12 : 1 & 1 : 1 \\
     & Time to Train, \si{\min}  & 39.53 $\pm$ 5.29 & 43.36 $\pm$ 3.90 & \textbf{36.56 $\pm$ 0.19} & \textbf{36.91 $\pm$ 0.07} & 37.48 $\pm$ 0.15 & 37.91 $\pm$ 0.73 \\ 
     & Validation Top 1, \% & 55.54 $\pm$ 0.34 & 53.89 $\pm$ 0.15 & \textbf{58.00 $\pm$ 0.17} & 70.49 $\pm$ 0.05 & 71.38 $\pm$ 0.07 & \textbf{71.47 $\pm$ 0.08} \\ 
\midrule
\multicolumn{8}{c}{Constant Global Batch Size} \\ \midrule
     &  & \multicolumn{3}{c|}{ViT B/16} & \multicolumn{3}{c}{ResNet-50}  \\
Nodes    &  & AB & AB - No Groups & Traditional DDP & AB & AB - No Groups & Traditional DDP  \\ \midrule
\multirow{4}{*}{4}     & Scaled Traffic, \si{\giga\byte / \s} & \textbf{2.49 $\pm$ 0.49} & 5.61 $\pm$ 0.16 & 8.93 $\pm$ 0.09 & \textbf{1.07 $\pm$ 0.01} & 2.69 $\pm$ 0.03 & 3.83 $\pm$ 0.01 \\
    & Compression   & 2.09 : 1 & \textbf{2.21 : 1} & 1 : 1 & \textbf{1.38 : 1} & 1.27 : 1 & 1 : 1 \\
     & Time to Train, \si{\min} & 259.44 $\pm$ 49.68 & 227.99 $\pm$ 5.29 & \textbf{224.97 $\pm$ 1.79} & \textbf{194.64 $\pm$ 0.91} & 195.76 $\pm$ 3.49 & 195.03 $\pm$ 0.74 \\ 
     & Validation Top 1, \% & \textbf{70.68 $\pm$ 0.56} & 68.66 $\pm$ 0.22 & 69.26 $\pm$ 0.18 & \textbf{74.75 $\pm$ 0.09} & 73.65 $\pm$ 0.13 & 73.74 $\pm$ 0.13 \\ \midrule
\multirow{4}{*}{8}     & Scaled Traffic, \si{\giga\byte / \s} & \textbf{4.89 $\pm$ 0.44} & 10.31 $\pm$ 0.32 & 13.09 $\pm$ 1.92 & \textbf{1.46 $\pm$ 0.18} & 3.98 $\pm$ 0.41 & 4.96 $\pm$ 0.09 \\
    & Compression   & \textbf{2.27 : 1} & 2.19 : 1 & 1 : 1 & \textbf{1.54 : 1} & 1.27 : 1 & 1 : 1 \\
     & Time to Train, \si{\min}  & \textbf{123.35 $\pm$ 5.81} & 124.85 $\pm$ 1.92 & 153.03 $\pm$ 17.77 & \textbf{135.98 $\pm$ 12.43} & 131.76 $\pm$ 12.52 & 139.82 $\pm$ 2.00 \\ 
     & Validation Top 1, \% & \textbf{69.66 $\pm$ 0.44} & 68.37 $\pm$ 0.15 & 69.38 $\pm$ 0.14 & \textbf{73.76 $\pm$ 0.19} & 73.60 $\pm$ 0.02 & 73.64 $\pm$ 0.09 \\ \midrule
\multirow{4}{*}{16}     & Scaled Traffic, \si{\giga\byte / \s} & \textbf{8.17 $\pm$ 0.17} & 17.14 $\pm$ 0.28 & \underline{25.00 $\pm$ 0.31} & \textbf{1.91 $\pm$ 0.27} & 5.57 $\pm$ 0.09 & 6.09 $\pm$ 0.46 \\
   & Compression  & \textbf{2.54 : 1} & 2.20 : 1 & 1 : 1 & \textbf{1.72 : 1} & 1.27 : 1 & 1 : 1 \\
     & Time to Train, \si{\min}  & 77.68 $\pm$ 9.18 & \textbf{74.26 $\pm$ 0.76} & 76.95 $\pm$ 0.87 & 111.46 $\pm$ 12.41 & \textbf{92.52 $\pm$ 1.78} & 113.00 $\pm$ 7.23 \\ 
     & Validation Top 1, \% & \textbf{69.19 $\pm$ 0.68} & 68.67 $\pm$ 0.34 & 68.79 $\pm$ 0.40 & 72.41 $\pm$ 0.04 & 73.54 $\pm$ 0.04 & \textbf{73.72 $\pm$ 0.03} \\ \midrule
\multirow{4}{*}{32}     & Scaled Traffic, \si{\giga\byte / \s} & \textbf{9.84 $\pm$ 0.07} & 21.81 $\pm$ 2.66 & \underline{25.00 $\pm$ 2.55} & \textbf{1.84 $\pm$ 0.42} & 5.70 $\pm$ 0.03 & 6.49 $\pm$ 1.85 \\
   & Compression   & \textbf{2.91 : 1} & 2.20 : 1 & 1 : 1 & \textbf{1.91 : 1} & 1.27 : 1 & 1 : 1 \\
     & Time to Train, \si{\min}  & \textbf{51.10 $\pm$ 0.29} & 81.41 $\pm$ 19.36 & 77.99 $\pm$ 18.29 & 110.09 $\pm$ 28.61 & \textbf{92.37 $\pm$ 3.35} & 121.55 $\pm$ 31.04 \\ 
     & Validation Top 1, \% & 66.13 $\pm$ 0.22 & 68.58 $\pm$ 0.57 & \textbf{68.99 $\pm$ 0.10} & 70.59 $\pm$ 0.08 & \textbf{73.73 $\pm$ 0.10} & 73.71 $\pm$ 0.04 \\ 
\bottomrule
\end{tabular}
\end{table*}
}

\usepackage{algpseudocode}

\def\BibTeX{{\rm B\kern-.05em{\sc i\kern-.025em b}\kern-.08em
    T\kern-.1667em\lower.7ex\hbox{E}\kern-.125emX}}
\begin{document}

\title{AB-Training: A Communication-Efficient Approach for Distributed Low-Rank Learning
}

\author{\IEEEauthorblockN{1\textsuperscript{st} Daniel Coquelin}
\IEEEauthorblockA{\textit{Scientific Computing Center (SCC)} \\
\textit{Karlsruhe Institute of Technology (KIT)}\\
Eggenstein-Leopoldshafen, Germany \\
daniel.coquelin@kit.edu}
\and
\IEEEauthorblockN{2\textsuperscript{nd} Katherina Fl\"ugel}
\IEEEauthorblockA{\textit{Scientific Computing Center (SCC)} \\
\textit{Karlsruhe Institute of Technology (KIT)}\\
Eggenstein-Leopoldshafen, Germany \\
katharina.fluegel@kit.edu}
\and
\IEEEauthorblockN{3\textsuperscript{rd} Marie Weiel}
\IEEEauthorblockA{\textit{Scientific Computing Center (SCC)} \\
\textit{Karlsruhe Institute of Technology (KIT)}\\
Eggenstein-Leopoldshafen, Germany \\
marie.weiel@kit.edu}
\and
\IEEEauthorblockN{4\textsuperscript{th} Nicholas Kiefer}
\IEEEauthorblockA{\textit{Scientific Computing Center (SCC)} \\
\textit{Karlsruhe Institute of Technology (KIT)}\\
Eggenstein-Leopoldshafen, Germany \\
nicholas.kiefer@kit.edu}
\and
\IEEEauthorblockN{5\textsuperscript{th} Muhammed \"Oz}
\IEEEauthorblockA{\textit{Scientific Computing Center (SCC)} \\
\textit{Karlsruhe Institute of Technology (KIT)}\\
Eggenstein-Leopoldshafen, Germany \\
muhammed.oez@kit.edu}
\and
\IEEEauthorblockN{6\textsuperscript{th} Charlotte Debus}
\IEEEauthorblockA{\textit{Scientific Computing Center (SCC)} \\
\textit{Karlsruhe Institute of Technology (KIT)}\\
Eggenstein-Leopoldshafen, Germany \\
charlotte.debus@kit.edu}
\and
\IEEEauthorblockN{7\textsuperscript{th} Achim Streit}
\IEEEauthorblockA{\textit{Scientific Computing Center (SCC)} \\
\textit{Karlsruhe Institute of Technology (KIT)}\\
Eggenstein-Leopoldshafen, Germany \\
achim.streit@kit.edu}
\and
\IEEEauthorblockN{8\textsuperscript{th} Markus G\"otz}
\IEEEauthorblockA{\textit{Scientific Computing Center (SCC)} \\
\textit{Karlsruhe Institute of Technology (KIT)}\\
Eggenstein-Leopoldshafen, Germany \\
markus.goetz@kit.edu}
}

\maketitle

\begin{abstract}
Communication bottlenecks severely hinder the scalability of distributed neural network training, particularly in high-performance computing (HPC) environments.
We introduce \emph{AB-training}, a novel data-parallel method that leverages low-rank representations and independent training groups to significantly reduce communication overhead.
Our experiments demonstrate an average reduction in network traffic of approximately 70.31\% across various scaling scenarios, increasing the training potential of communication-constrained systems and accelerating convergence at scale.
AB-training also exhibits a pronounced regularization effect at smaller scales, leading to improved generalization while maintaining or even reducing training time.
We achieve a remarkable 44.14 : 1 compression ratio on VGG16 trained on CIFAR-10 with minimal accuracy loss, and outperform traditional data parallel training by 1.55\% on ResNet-50 trained on ImageNet-2012.
While AB-training is promising, our findings also reveal that large batch effects persist even in low-rank regimes, underscoring the need for further research into optimized update mechanisms for massively distributed training.
\end{abstract}


\section{Introduction}

The relentless pursuit of predictive performance in machine learning has led to neural networks of unprecedented scale and complexity. 
These massive state-of-the-art models excel in tasks ranging from image classification and natural language processing to scientific simulations. 
However, this performance comes at a steep computational cost, as training large models often demands large datasets.
The sizes of modern models and datasets necessitates the usage of distributed training approaches.

Data parallelism (DP) is the most common distributed training technique for neural networks~\cite{tal2019survey}.
It replicates the model across multiple devices, each of which processes a subset of the data, known as the local batch, before aggregating the gradients and updating model parameters. 
A persistent challenge in data parallel (DP) training is the communication bottleneck created by synchronizing large model representations across compute nodes~\cite{niu_hogwild_2011}. 

As the accelerators of neural networks increase in computing power and on-board memory, models finish the forward-backward pass faster.
In the standard implementations of DP training, the model gradients are eagerly synchronized once they are calculated during the backward pass. 
As models get larger and accelerators get faster, this synchronization happens more frequently and with more data.
To solve this, one must either increase the available interconnect bandwidth or reduce the rate at which packets are sent to other nodes.

This bottleneck severely limits scalability, particularly on distributed-memory computing systems with potential network constraints.
Furthermore, DP training increases the batch size used to calculate the gradients themselves which, at large scales, reduces the generalizablity of the network~\cite{goyal_accurate_2018}.
These large batch effects pose an additional challenge for maintaining predictive performance, as accuracy degrades once the global batch size crosses a critical threshold.

To address these challenges, we propose a novel low-rank training method based on the singular value decomposition (SVD) and hierarchical training specifically designed to reduce the interconnect traffic in distributed settings. 
Our method leverages the fact that low-rank representations can reduce the amount of data transferred~\cite{vogels_powersgd_2019}, as well as the observation that low-rank representations can promote regularization~\cite{coquelin_harnessing_2024}.
Key contributions of our work include:
\begin{itemize}
    \item Through the usage of independent training groups and low-rank representations, we reduce the network traffic during distributed training by an average of70.31\% as compared to traditional synchronous DP training. 
    \item Our method demonstrates improved regularization effects at smaller scales, particularly on the Vision Transformer (ViT), leading to better predictive performance on unseen data. 
    \item In an ideal scenario, AB training achieves a compression ratio of 44.14 : 1 for the VGG16~\cite{liu2015vgg} model when trained on CIFAR-10~\cite{krizhevsky2009cifar} while maintaining a competitive accuracy. In more standard scenarios, it achieved compression values ranging from 1.19 : 1 to 2.54 : 1 when exceeding the accuracy of traditional DP training.
\end{itemize}
Our work highlights promising directions for how to harness the potential of low-rank training in large-scale scenarios, with implications for both distributed-memory computing and the broader ML community.
We provide an open-source implementation of our method, as well as the configurations to reproduce the experiments shown\footnote{Will be made available upon publication}.

\section{Related Work}

\subsection{Distributed Training of Neural Networks}

The computational demands of training large neural networks have driven the development of distributed training paradigms, with data parallelism as a cornerstone approach.  
In a traditional data parallel (DP) training scheme, the model is replicated across multiple devices, each processing a unique subset of the training data independently.
The data subset that one model instance processes is the local batch, while the global batch is the set of all local batches.
After each forward-backward pass, gradients are aggregated across all model instances before the model is updated by the optimizer. 

As the number of compute resources and the model size increase, communication overheads incurred during gradient synchronization become a significant constraint. 
Other parallelization strategies, such as model parallelism, introduce alternative ways to distribute training across devices, but come with their own communication and load-balancing challenges.

Several techniques have been proposed to mitigate the communication bottleneck in data parallel training, including gradient accumulation, topology-aware communication patterns, asynchronous methods, and gradient compression~\cite{vogels_powersgd_2019,abrahamyan2022compression}.
Methods like H-SGD~\cite{lin_dont_2020} and DASO~\cite{coquelin_accelerating_2022} utilize localized synchronization within smaller groups between global updates, reducing communication costs by leveraging the system's network topology.
Asynchronous approaches typically make use of a parameter server for aggregating gradients~\cite{niu_hogwild_2011} and often require extensive hyperparameter tuning~\cite{tal2019survey}. 
Despite these advancements, the scalability of data parallelism remains constrained by communication bottlenecks and large batch effects, where increasing the global batch size negatively impacts generalization performance.

\subsection{Low-Rank Neural Network Training}

A promising avenue for reducing both the memory footprint and computational cost of neural networks lies in low-rank training.  
Singular Value Decomposition (SVD) plays a crucial role in this domain. 
By factoring a neural network's weight matrix $W$ into three matrices (an orthogonal basis $U$, a diagonal matrix of singular values $\Sigma$, and an orthogonal basis $V^T$) and retaining only the $k$ largest singular values, a low-rank approximation with significantly fewer parameters can be obtained given a sufficiently large $k$.
However, aggressive compression can lead to a loss of approximation quality.

Several methods leverage SVD for compressing neural networks during training. 
Some train directly on the $U$, $\Sigma$, and $V$ matrices~\cite{schotthofer_low-rank_2022,coquelin_harnessing_2024}, while others create different representations using the SVD and training on them~\cite{Liebenwein2021compressing,yang2020learning}.
Both approaches can offer regularization-like benefits, potentially improving generalization~\cite{winata2020lowranktrans,cahyawijaya2021greenformer,phan2020stablecnn}.
LoRA~\cite{hu2021lora}, and methods similar to and inspired by it, utilize low-rank representations alongside the frozen full-rank weights for fine-tuning.

Gradient compression using SVD has found application in federated learning scenarios~\cite{wang_fedsvd_2023}, where bandwidth constraints necessitate reduced communication. 
While these techniques address computational challenges in neural network training, their integration within distributed environments to specifically optimize communication remains an active area of research.  
More precisely, existing approaches simply train low-rank models in parallel, but do not make any efforts to use the distributed system to their advantage~\cite{coquelin_harnessing_2024,Wang2023CuttlefishLM,Wang2021PufferfishCM}.

\subsection{Pruning}

Pruning aims to reduce model size by removing irrelevant parameters.
It is the primary competition to low-rank methods for reducing the size of neural networks.
One well-known method is Iterative Magnitude Pruning (IMP), which involves repeatedly training a network, removing small-magnitude weights, and retraining from the initial state\cite{wimmer2023surveydecomp}.
While this process can result in highly compressed models., unstructured pruning methods like IMP often lead to irregular sparsity patterns, making it difficult to leverage specialized hardware optimized for sparse computation.
Structured pruning techniques target specific components of the network.
For example, in convolutional layers removing filters \cite{singh_play_2019,he_soft_2018,luo_thinet_2017} or channels \cite{lin2020abcpruning,he_channel_2017,chang2023icpprune} can maintain traditional dense parameter layouts to make use of efficient, pre-existing acceleration methods.

\section{Methodology}

Data parallel training of neural networks is constrained by the the communication bottleneck caused by synchronizing massive model representations across compute nodes and large batch effects, a phenomenon where accuracy degrades at large global batch sizes.
In an attempt to tackle these challenges, our proposed training method integrates low-rank weight representations within a hierarchical data parallel framework.
It leverages the reduced network capacity stemming from topologically informed communication patterns and low-rank training while also escaping local minima using group training.

Working with large batches can have unintended consequences.
Consider the standard batch stochastic gradient descent (SGD) optimization step
\begin{equation}
    W_t = W_{t-1} - \frac{\eta}{B}\sum_{i=1}^{B}\nabla Q_i\left(W_{t-1}\right),
\end{equation}
where $W_t$ is a parameter at time $t$ in training, $\eta$ is the learning rate, $N$ is the batch size, and $Q_i\left(W\right)$ is the value of the loss function for the $i$-th data element.
It is easily shown that for a given iteration $k > 0$, the weights are simply the sum of all the previous update steps or
\begin{equation}
    W_k = W_{0} - \frac{\eta}{B}\sum_{s=1}^{k-1}
    \sum_{i=1}^{B}\nabla Q_i\left(W_{s}\right)
\end{equation}
With a larger batch size, $B$, the magnitude of the gradients shrinks.
This removes the randomness inherent to SGD's success and makes the network more prone to getting stuck in local minima~\cite{Chaudhari_2019_entropy,smith2020noise}.

To mitigate this effect, we draw motivation from the fact that the noise inherent to the gradients in minibatch SGD is an important factor in why it generalizes better than large batch gradient descent~\cite{smith2020noise}. 
During training we divide the workers into small sub-groups; each of which train the model independently for a number of steps.
By dividing the DP model replicas into subgroups to be trained on independent data subsets, we effectively form an ensemble of models. 
Individually, these groups benefit from the improved generalization of minibatch SGD by using the noisier gradients to explore of the loss landscape to find minima which would otherwise be missed by large-batch SGD.
Additionally, training in independent groups heavily reduces the network traffic.

To further reduce the amount of data communicated during training, we utilize SVD to decompose weight matrices into low-rank representations.
We represent the weights of a given layer as 
\begin{equation}
    W_{m \times n} = A_{m \times k}B_{k \times n} = \left( U_{m \times k} \Sigma_{k \times k}^{1/2} \right)\left( \Sigma_{k \times k}^{1/2} V_{k \times n}^\top \right)
    \label{eq:ab}
\end{equation}
where $U$, $\Sigma$, and $V$ are determined by the SVD of $W$ and $k < \texttt{min}\left(m, n\right)$.
In the case that the weight matrix has more than two dimensions, the trailing dimensions are flattened to form a two-dimensional representation to be used as $W$.
If the shape of the matrix $W$ is unfavorable ($n > m$) we transpose $W$ before the decomposition.

One of the most obvious methods for training a low-rank network would be to train directly on the $\Sigma$ matrix.
However, it is not beneficial to have any trainable weights set to zero during training as these connections are effectively dead during backpropagation.
Since the off-diagonal values of $\Sigma$ (as dictated by SVD) are zero, training on only $\Sigma$ will have the effect of freezing the orthogonal vectors of $U$ and $V$.
To allow the network to modify the orthogonal vectors and to maintain a dense representation, we decompose the network as shown in \Cref{eq:ab}.
Interestingly, training a network using low-rank representations has the added benefit of reducing overfitting, a common problem where the performance on the training dataset is much higher than on the validation or test datasets~\cite{yang2020learning,hu_low_2021}.

During the independent training phase, half of the sub-groups train $A$ while the others train $B$; in this phase the other component remains fixed. 
This ensures that at least a portion of the orthogonal vectors inherent to $W$ are fixed in time, reducing the chance that the groups greatly diverge and increases the gradient noise due to the smaller batch size.
With this method, the groups can more easily learn to modify the full low-rank representations of the network's parameters.

At the end of the independent training phase, we average $A$ and $B$ across all workers to merge the independently trained models.
This merging strategy has seen success in federated learning~\cite{giuseppi_weighted_2022}, a use-case which typically deals with models trained independently on biased datasets.
As we use unbiased batches during the independent training phase, we expect the negative aspects of this average to be minimal.

Our \emph{AB} training procedure consists of the following phases:
\begin{itemize}
    \item \textbf{Full Rank DP Warmup}: Initial training with full-rank data parallelism allows the weights to move rapidly away from their random initialization. To avoid early divergence, we employ a traditional, full-rank data parallel warmup phase for a given number of iterations (\texttt{warmupSteps}). 
    \item \textbf{Independent AB decomposition}: 
    Each model instance independently computes an AB decomposition of its weight matrices as per \Cref{eq:ab}. 
    This AB decomposition allows us to represent the weight matrices more compactly, reducing the amount of data to be communicated during training.
    The rank of this approximation is determined by a user-provided hyperparameter (\texttt{sigmaCutoff}) and the largest singular value. 
    Values smaller than the largest singular value times \texttt{sigmaCutoff} are removed.
    \item \textbf{Group training}: 
    Half of the independent groups will train the $A$ matrix, denoted as the $A$ groups, while the other groups will train $B$, the $B$ groups.
    For a given number of iterations (\texttt{numABSteps}), the $A$ group only trains the $A$ matrices and the $B$ group only trains the $B$ matrices, the matrices not being trained are frozen. 
    \item \textbf{Synchronization and update}: 
    We average both the trained and frozen $A$ and $B$ matrices to synchronize the model instances across all processes. 
    Afterwards, we reconstruct the full-rank weight matrices from the updated $A$ and $B$ components. 
    \item \textbf{Full-rank rebound}: 
    We then train the full-rank network with traditional data parallelism for a number of iterations (\texttt{fullRankReboundSteps}). 
    This phase promotes convergence and helps mitigate potential accuracy degradation observed in low-rank training scenarios.
\end{itemize}

\begin{figure}[tb]
    \centering
    \includegraphics[width=0.9\linewidth]{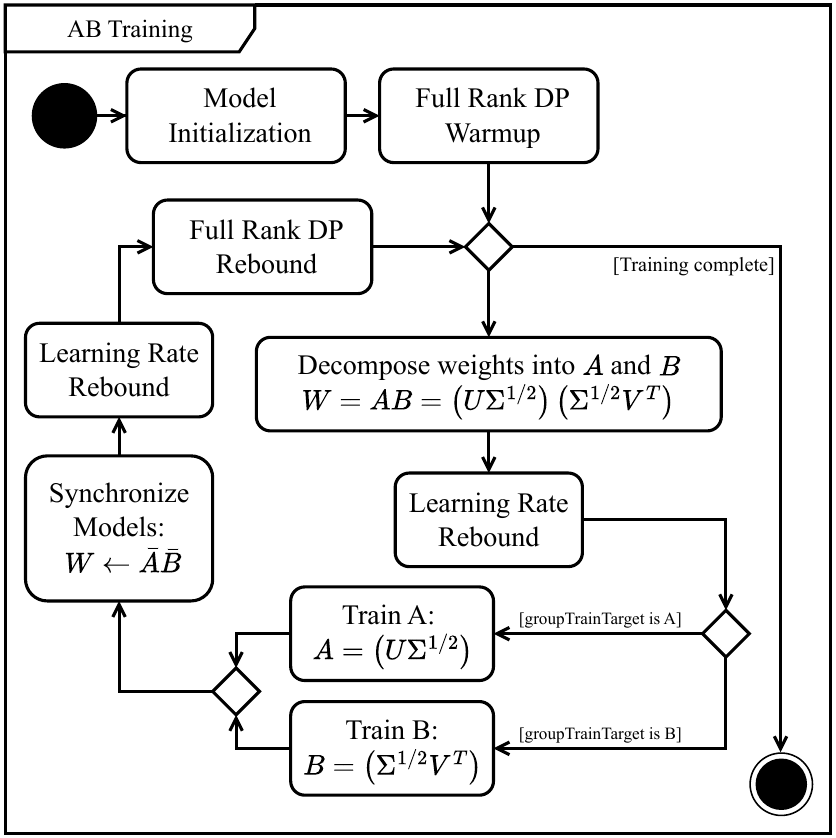}
    \caption{A UML diagram of the AB training procedure.}
    \label{fig:ab-training}
\end{figure}

A formalized algorithm is shown in \Cref{algo:main} and a method diagram is shown in \Cref{fig:ab-training}.

\begin{algorithm}
\DontPrintSemicolon
\SetKwInOut{Input}{Input}
\SetKwInOut{Output}{Output}
\Input{Model, $M$ , training data, $numTrainingSteps$, hyperparameters ($warmupSteps$, $numABSteps$, $fullRankReboundSteps$)}

\If{procId $\leq$ worldSize / 2}{$groupTrainTarget \gets A$}
\Else{$groupTrainTarget \gets B$ }


\For{$i \gets 1$ to warmupSteps}{ 
  traditionalFullRankDPTraining()
}
\Repeat {numTrainingSteps = completed steps}{
  \ForEach{$W$ \textnormal{in} $M$}{
    $A, B \gets \texttt{abDecomposition}(W)$ \hfill\Comment{Eq. \ref{eq:ab}} \;
    removeSmallSingularValues()
  }
  startLearningRateRebound()\;
  $groupTrainTarget$.setTrainable(True)\;
  
  \For{$i \gets 1$ to numABSteps}{
      independentSubgroupTraining()
  }
  $A\gets \texttt{allReduce}(A/worldSize$)\;
  $B\gets \texttt{allReduce}(B/worldSize$)\;
  \ForEach{$W$ \textnormal{in} $M$}{
    \textnormal{$W\gets AB$}
  }
  startLearningRateRebound()\;
  \For{$i \gets 1$ to fullRankReboundSteps}{
      traditionalFullRankDPTraining()
  }
}
\caption{The AB training method. $W$ is a parameter of the input model $M$ and $worldSize$ is the number of workers used for traditional DP training.} 
\label{algo:main}
\end{algorithm}

Popular neural network optimizers like AdamW~\cite{loshchilov2018adamw} rely on second-order derivative approximations. 
When we change the network's structure, e.g., between low-rank and full-rank representations, the optimizer states used in these approximations become incompatible due to their dimensionally. 
To address this, we use a \emph{learning rate rebound} strategy.
This involves reducing the learning rate to near-zero and gradually increasing it back to its scheduled value over a user-provided number of steps, allowing the optimizer to adjust to the new parameter representations.
We do a learning rate rebound whenever the network's parameters change shape.

\section{Experimental Evaluation}

To evaluate the effectiveness of our AB training approach, we conducted experiments on well-studied neural network architectures. 
Our primary objectives were to demonstrate significant reductions in interconnect traffic, potential regularization benefits, scalability, and achievable compression during training.
We focused on the ResNet-50~\cite{he_deep_2016} and Vision Transformer~\cite{dosovitskiy_image_2021} B/16 models trained on the ImageNet-2012 dataset~\cite{krizhevsky_imagenet_2012} for image classification as well as VGG16~\cite{liu2015vgg} trained on CIFAR-10~\cite{krizhevsky2009cifar}.
We compare our results against traditional full-rank DP training as implemented by PyTorch's \texttt{DistributedDataParallel} (DDP) class, referred to as `Traditional DDP' or `Trad. DDP', using the same hyperparameters (HPs).
In these experiments, the $B$ matrix is trained during the group phase using traditional DP training.
The ViT HPs are those used in the original source, while the ResNet-50 parameters utilize the same learning rate scheduler as the ViT and the default values for the AdamW~\cite{loshchilov2018adamw} optimizer.
The VGG16 HPs are as shown in~\cite{coquelin_harnessing_2024}.

We analyze our training method using two scaling strategies: 
constant \textit{local} batch size scaling and constant \textit{global} batch size scaling.
In constant local batch size scaling, the global batch size increases with the number of GPUs.
In constant global batch size scaling, we maintain a constant global batch size while proportionally decreasing the local batch size as we increase the number of GPUs. 
This emphasizes the interconnect traffic reduction achieved by our method.
To disentangle the effects of training with independent groups and those of low-rank training, we show measurements for AB training with and without groups.
For the no-group measurements, the workers train the $B$ matrix with traditional DP methods and denoted as `AB - No Groups'.
At a fixed scale, we compare AB training with and without groups to other pruning and low-rank methods.
Measurements represent the average of three runs, each of which utilizes a different random seed.
Models are initialized using orthogonal initialization~\cite{Hu2020Provable}.

\subsection{Computational Environment}

We ran all experiments on a distributed-memory, parallel hybrid supercomputer. 
Each compute node is equipped with two 38-core Intel Xeon Platinum 8368 processors at \SI{2.4}{\giga\hertz} base and \SI{3.4}{\giga\hertz} maximum turbo frequency, \SI{512}{\giga\byte} local memory, a local \SI{960}{\giga\byte} NVMe SSD disk, two network adapters, and four NVIDIA A100-40 GPUs with \SI{40}{\giga\byte} memory connected via NVLink. 
Inter-node communication uses a low-latency, non-blocking NVIDIA Mellanox InfiniBand 4X HDR interconnect with \SI{200}{\giga\bit/\second} per port. 
All experiments used Python 3.10.6 with \texttt{CUDA}-enabled \texttt{PyTorch} 2.0.0~\cite{paszke2018pytorch}.

\subsection{Datasets and Models}

We used the ImageNet-2012 dataset, which contains 1.2 million images, for our scaling experiments. 
Basic image augmentation was applied, including normalization, random resizing, random cropping, and a random flip.  
We trained two models on this dataset: ResNet-50~\cite{he_deep_2016} and Vision Transformer B/16 (ViT)~\cite{dosovitskiy_image_2021}, chosen for their widespread use and distinct architectures.
We also trained VGG16 on the CIFAR-10 dataset.
Basic image augmentation was applied in the same way as for the ImageNet experiments.
All models were trained with the AdamW~\cite{loshchilov_decoupled_2019} optimizer.
Other training hyperparameters can be found in Appendix 1.

\subsection{Hyperparameter Considerations}

As with any method, hyperparameter tuning is essential to performance.
Key hyperparameters (HPs) of AB training include the warmup and full-rank rebound phase durations, the number of AB training iterations, the AB decomposition rank-reduction parameter, and the frequency of SVD and synchronization steps.
Additionally, the latter will influence the trade-offs between regularization and communication efficiency.

A hyperparameter search on CIFAR-100~\cite{krizhevsky2009cifar} with a minified Vision Transformer model (patch size of eight instead of 16, six heads instead of twelve, depth of six instead of twelve) was conducted using \texttt{propulate}~\cite{taubert_massively_2023} to determine the ideal settings for AB-Training's hyperparameters.
Using Weights and Biases~\cite{wandb}, we measures the importance of these parameters via a Bayesian analysis of the completed runs.
We find that the most important of these was the \texttt{numABStesps}.
Controlling how many steps to train in low rank was crucial.
However, the correlation of this variable to the final top-1 accuracy was only 0.225, indicating that there is a sweet spot between too many and too few steps.
The number of warmup steps was also found to be quite important, with short warmups performing worse.
The search and our analysis of the results suggested the following guidelines which are used for all experiments:
\begin{itemize}
    \item \texttt{warmupSteps}: 20\% of the total training steps
    \item \texttt{numABSteps}: 3.33\% of the total training steps
    \item \texttt{fullRankReboundSteps}: $0.25 \cdot\texttt{numABSteps}$
    \item Learning rate rebound steps: $0.5 \cdot\texttt{numABSteps}$
\end{itemize}

\subsection{Constant Local Batch Size Scaling}

In this experiment, we maintain a fixed \textit{local} batch size while increasing the available compute resources and decreasing the number of training steps per epoch, resulting in an increasing global batch size but less training iterations.
For example, if the global batch size for a two-node (eight-GPU) run is 2,048 and it is trained for 600 iterations in each epoch, the corresponding four-node (16-GPU) run would have a global batch size of 4,096 and be trained for 300 iterations each epoch.
This strategy investigates the communication efficiency gains achievable as we increase the available compute resources, even as the potential negative impacts of large batch sizes become more pronounced.
It is the most likely method by which a non-expert user will scale their training.

This experiment tests how well AB training reduces communication requirements and if the independent training groups can effectively train networks independently. 
We expect that if the independent training groups make similar updates, the averaged model will maintain accuracy and exhibit increased compression.
However, significant divergence in individual groups could lead to lower accuracy and reduced compression as the individual updates conflict.
If this is the case, then AB training without independent groups will begin to outperform AB training with groups.

\begin{figure}[tb]
    \centering
    \begin{subfigure}{\linewidth}
        \centering
        \includegraphics[width=\linewidth]{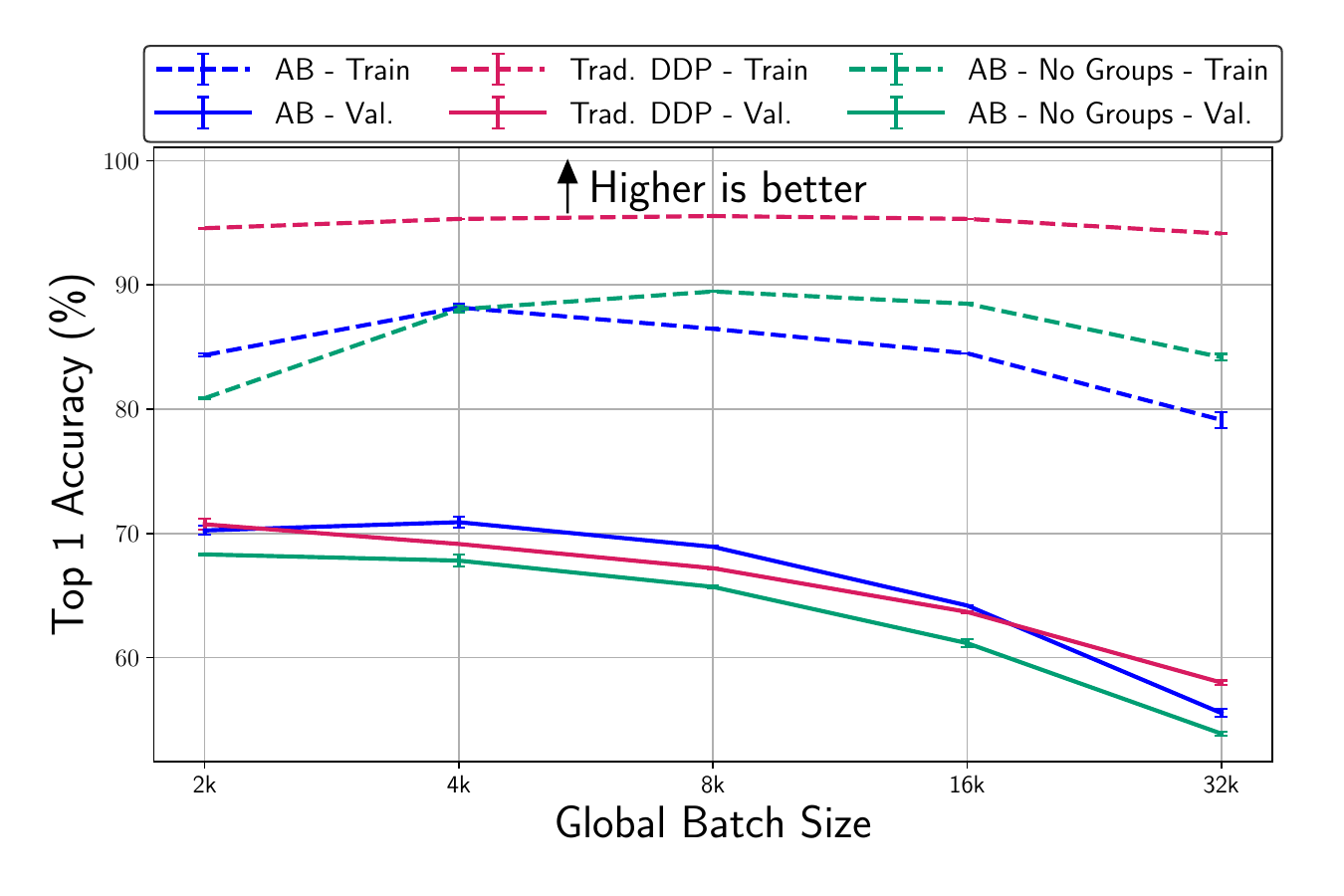}%
        \caption{Vision Transformer B/16}%
        \label{fig:vit-strong-top1}%
    \end{subfigure}
    \begin{subfigure}{\linewidth}
        \centering
        \includegraphics[width=\linewidth]{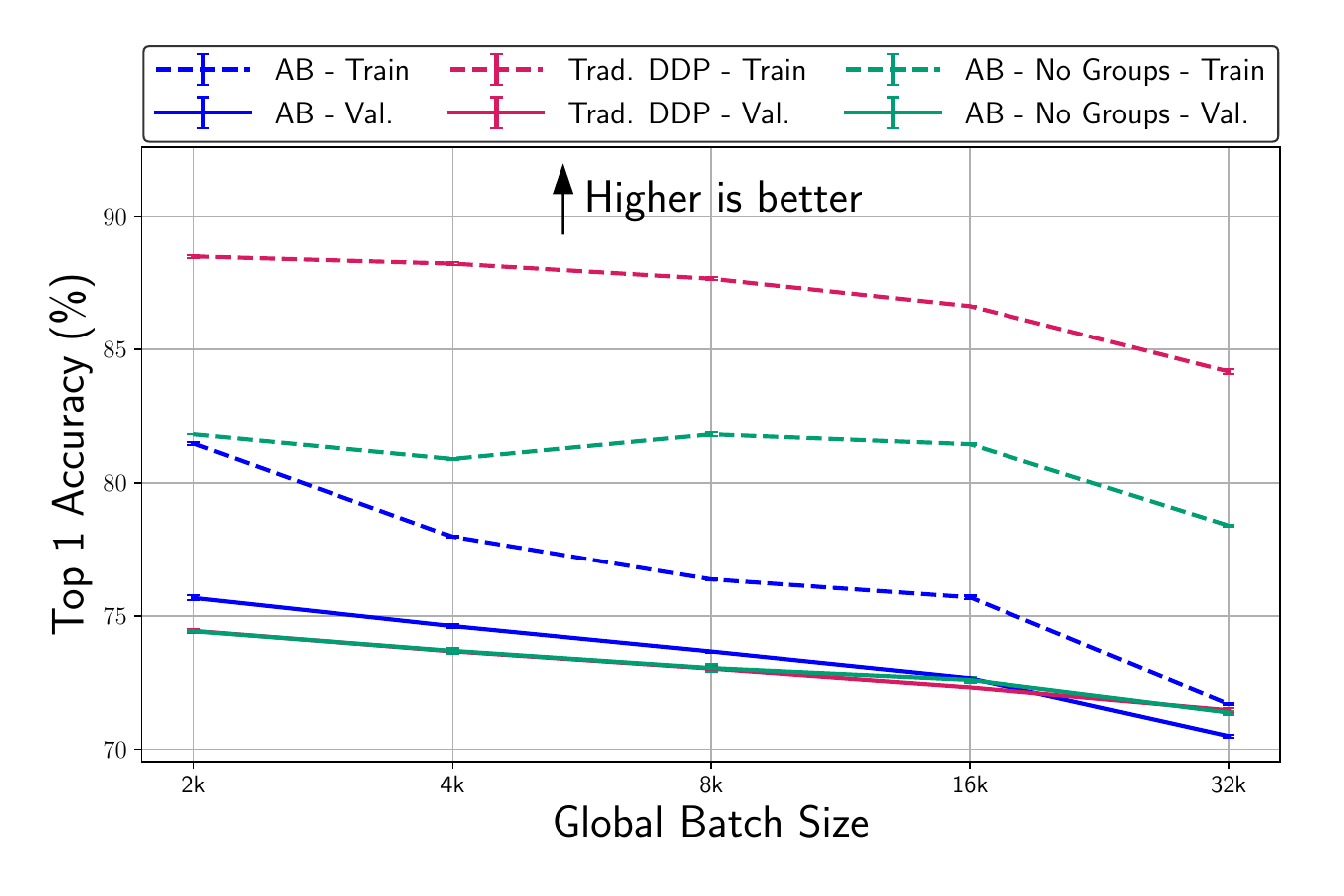}%
        \caption{ResNet-50}%
        \label{fig:rn50-strong-top1}%
    \end{subfigure}%
    \caption{Highest top 1 accuracy measurements for each training run on ImageNet-2012 for two network architectures with a constant \textit{local} batch size of 256. Global batch sizes range from 2,048 to 32,768 in powers of 2. Error are plotted, though not always visible.}
    \label{fig:top1s-strong}
\end{figure}

\Cref{fig:top1s-strong} shows the highest top 1 accuracy achieved by each model during training.
The compression values for the trained models are shown in \Cref{fig:compress-strong}.

\subsection{Constant Global Batch Size Scaling}

In this experiment, we maintain a fixed \textit{global} batch size while scaling compute resources. 
This contrasts with the standard approach of keeping local batch sizes constant, emphasizing the interconnect traffic reductions achieved by AB training.  
This will reduce the number of computations per process per forward-backward pass.
With a fixed global batch size of 4,096, a four-node run would have a local batch size of 256, whereas an eight-node run would have a local batch size of 128.

\begin{figure}[tb]
    \centering
    \begin{subfigure}{\linewidth}
        \centering
        \includegraphics[width=\linewidth]{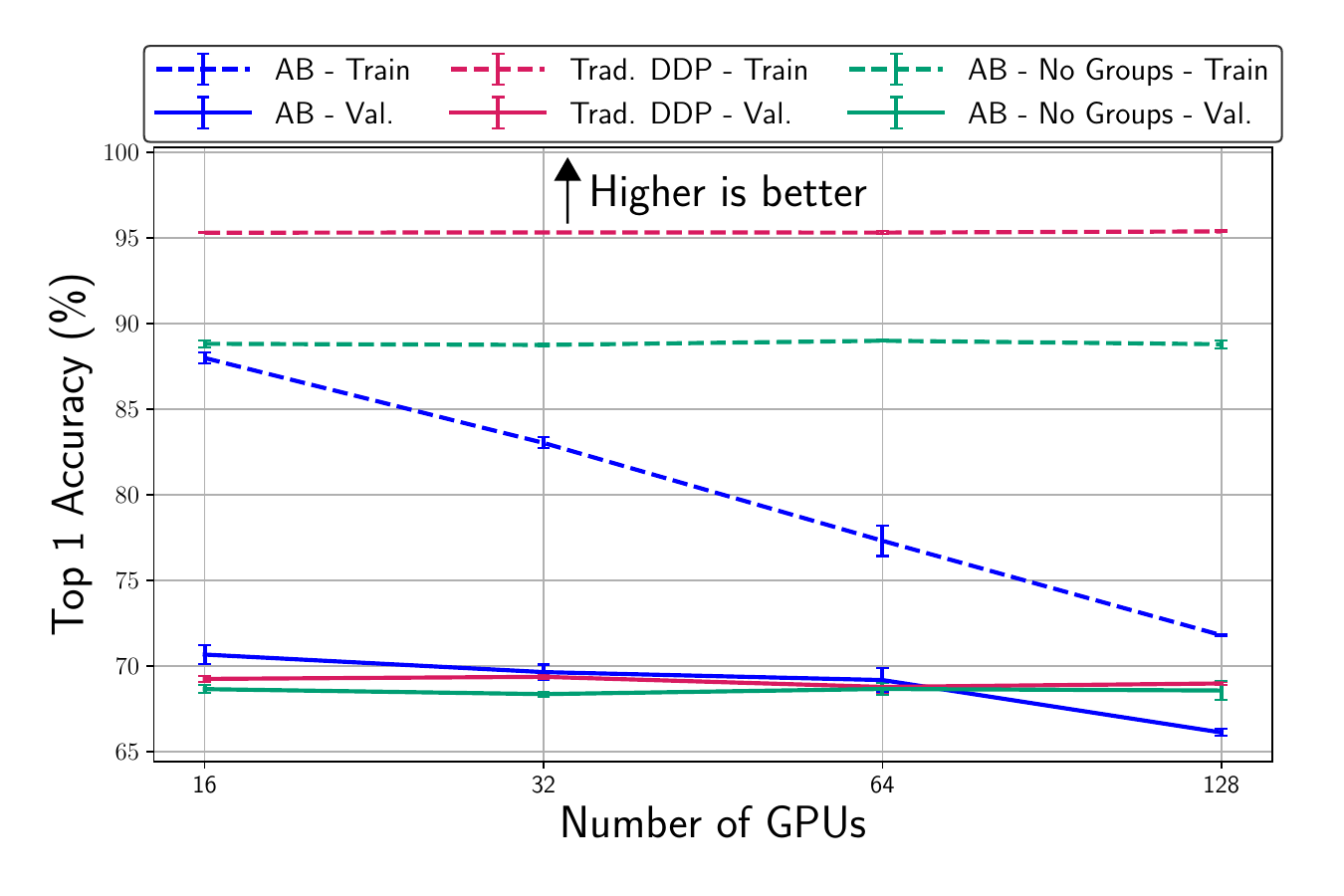}%
        \caption{Vision Transformer B/16}%
        \label{fig:vit-weak-top1}%
    \end{subfigure}
    \begin{subfigure}{\linewidth}
        \centering
        \includegraphics[width=\linewidth]{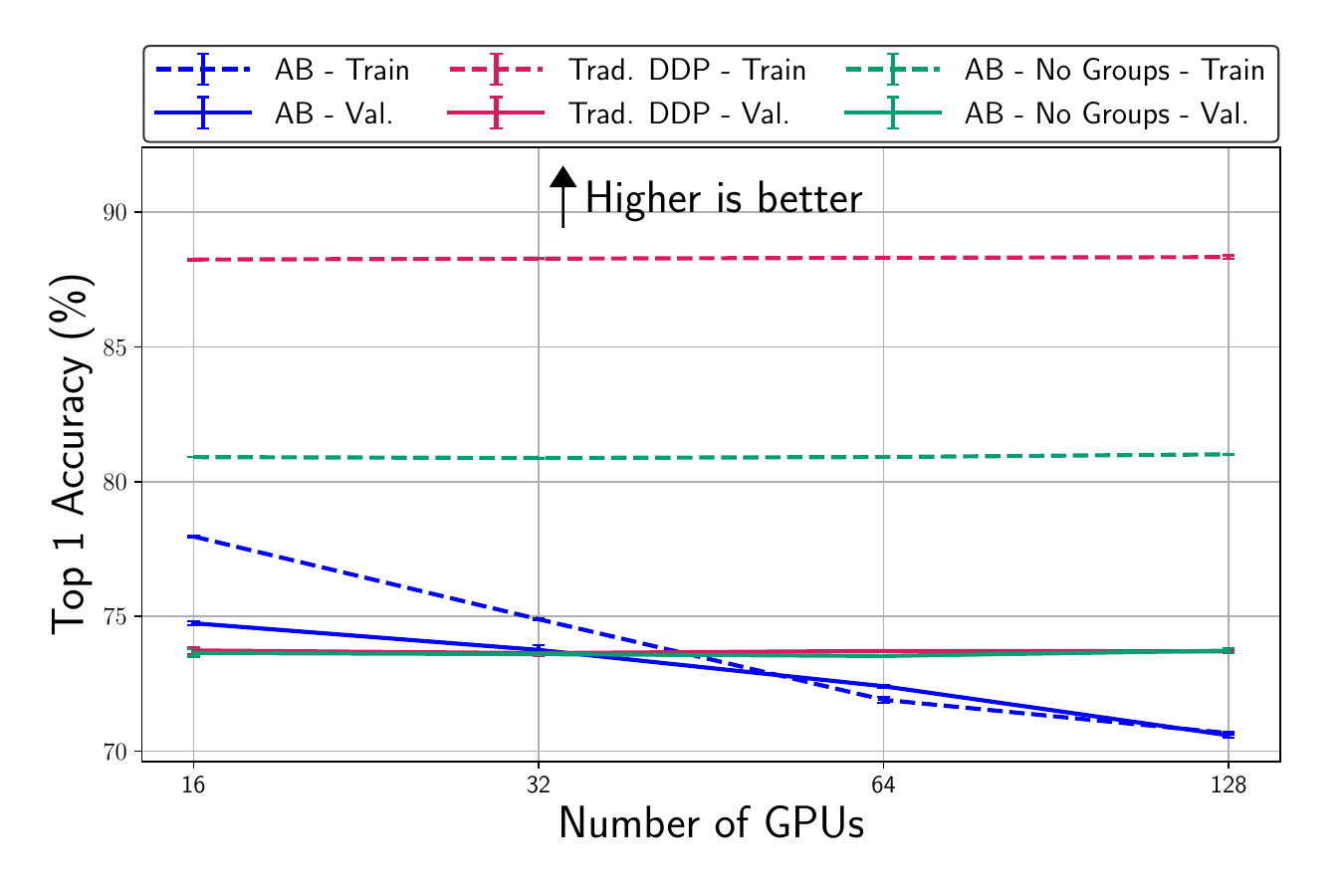}%
        \caption{ResNet-50}%
        \label{fig:rn50-weak-top1}%
    \end{subfigure}%
    \caption{Highest top 1 accuracy measurements for each training run on ImageNet-2012 for two network architectures with a constant \textit{global} batch size of 4,096. Error are plotted, though not always visible.}
    \label{fig:top1s-weak}
\end{figure}

\begin{figure}[tb]
    \centering
    \begin{subfigure}{\linewidth}
        \centering
        \includegraphics[width=\linewidth]{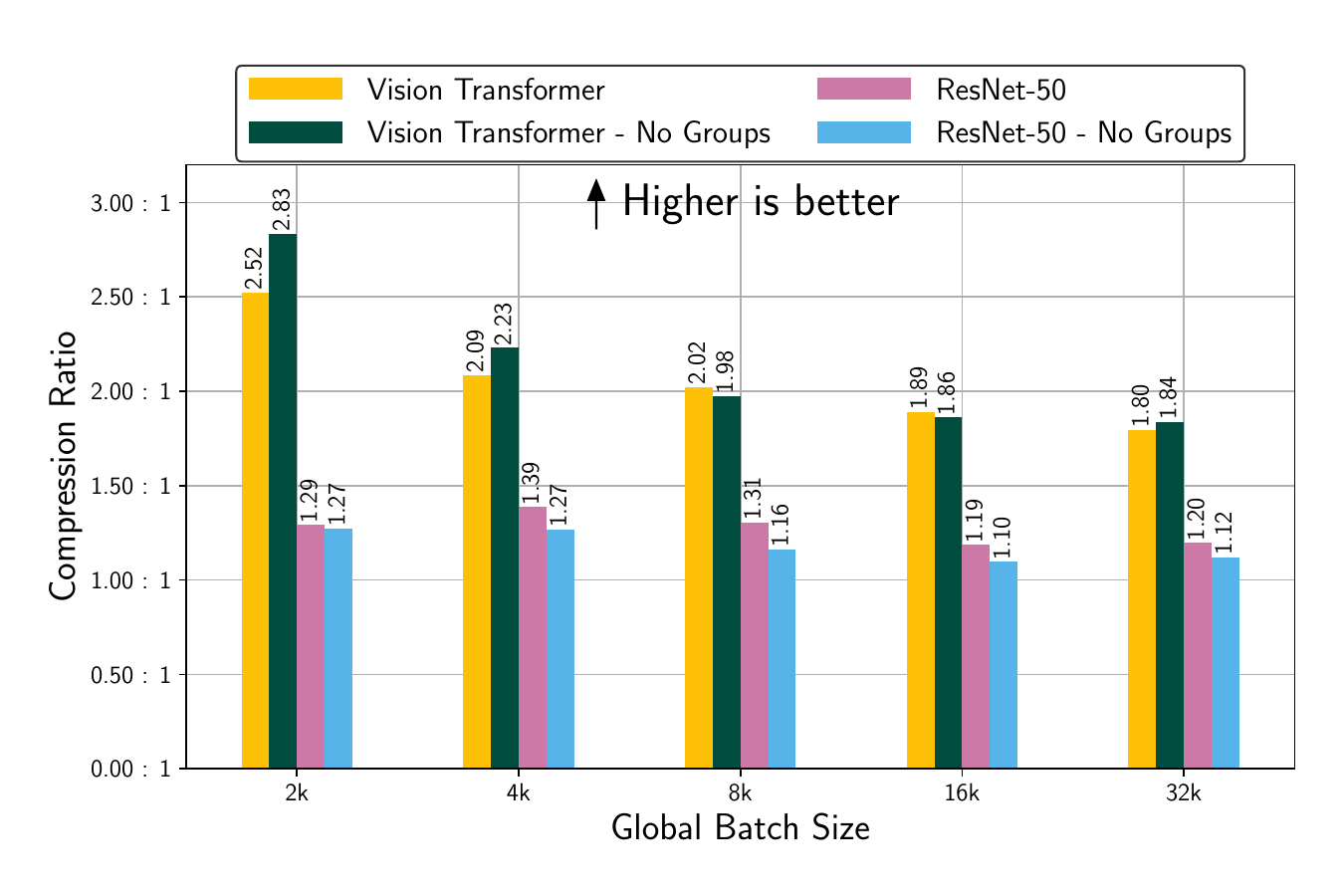}
        \caption{Constant \textit{local} batch size of 256.}
        \label{fig:compress-strong}
    \end{subfigure}
    \begin{subfigure}{\linewidth}
        \centering
        \includegraphics[width=\linewidth]{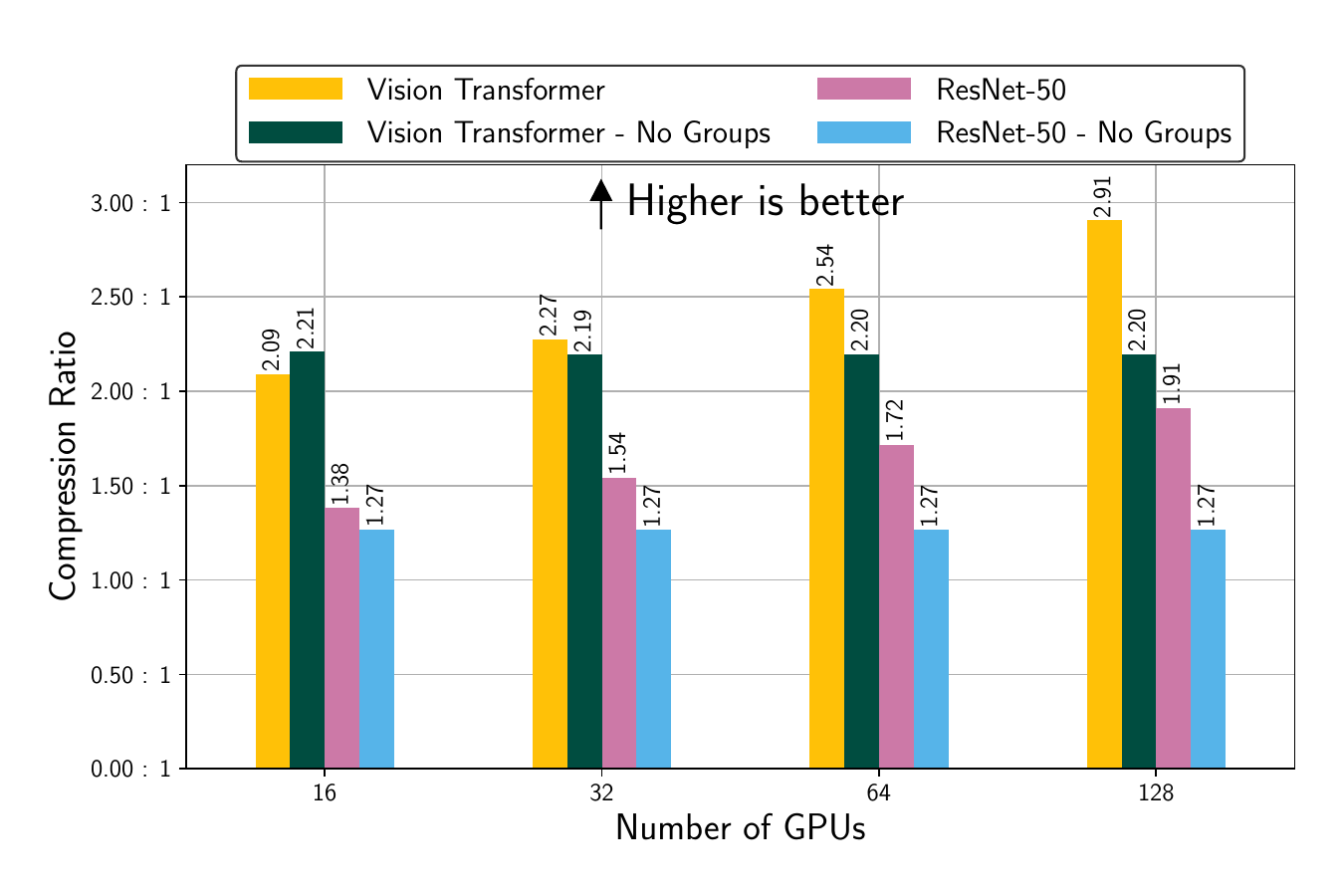}
        \caption{Constant \textit{global} batch size of 4,096.}
        \label{fig:compress-weak}
    \end{subfigure}%
    \caption{Compression ratios for AB training with and without groups and a traditional DP baseline on ImageNet-2012 for two network architectures}
    \label{fig:compress}
\end{figure}

\begin{figure}[tb]
    \centering
    \begin{subfigure}{\linewidth}
        \centering
        \includegraphics[width=\linewidth]{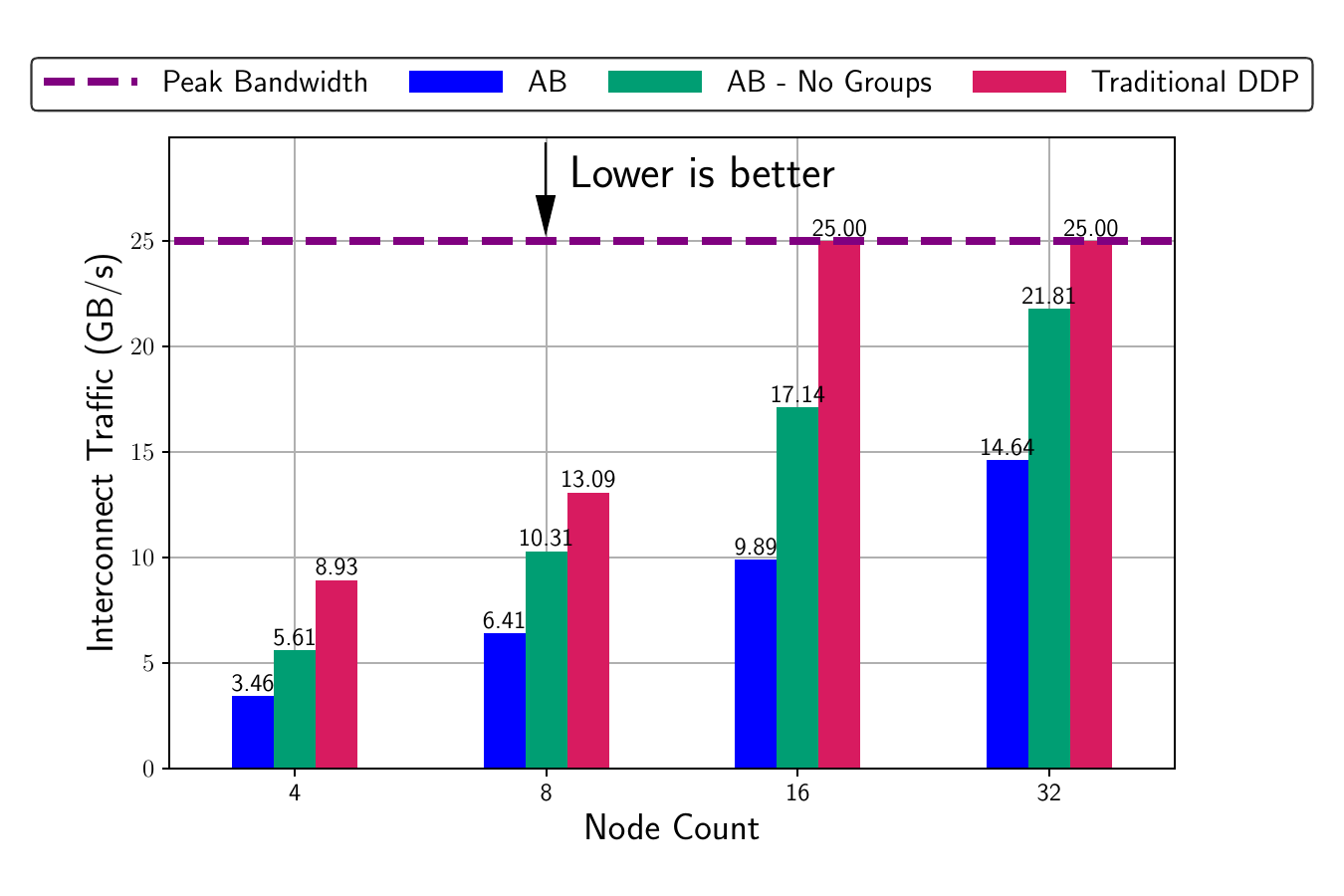}%
        \caption{Scaled interconnect traffic. The `Peak Bandwidth' is that of each port of each node's Infiniband card.}%
        \label{fig:weak-traffic}%
    \end{subfigure}
    \begin{subfigure}{\linewidth}
        \centering
        \includegraphics[width=\linewidth]{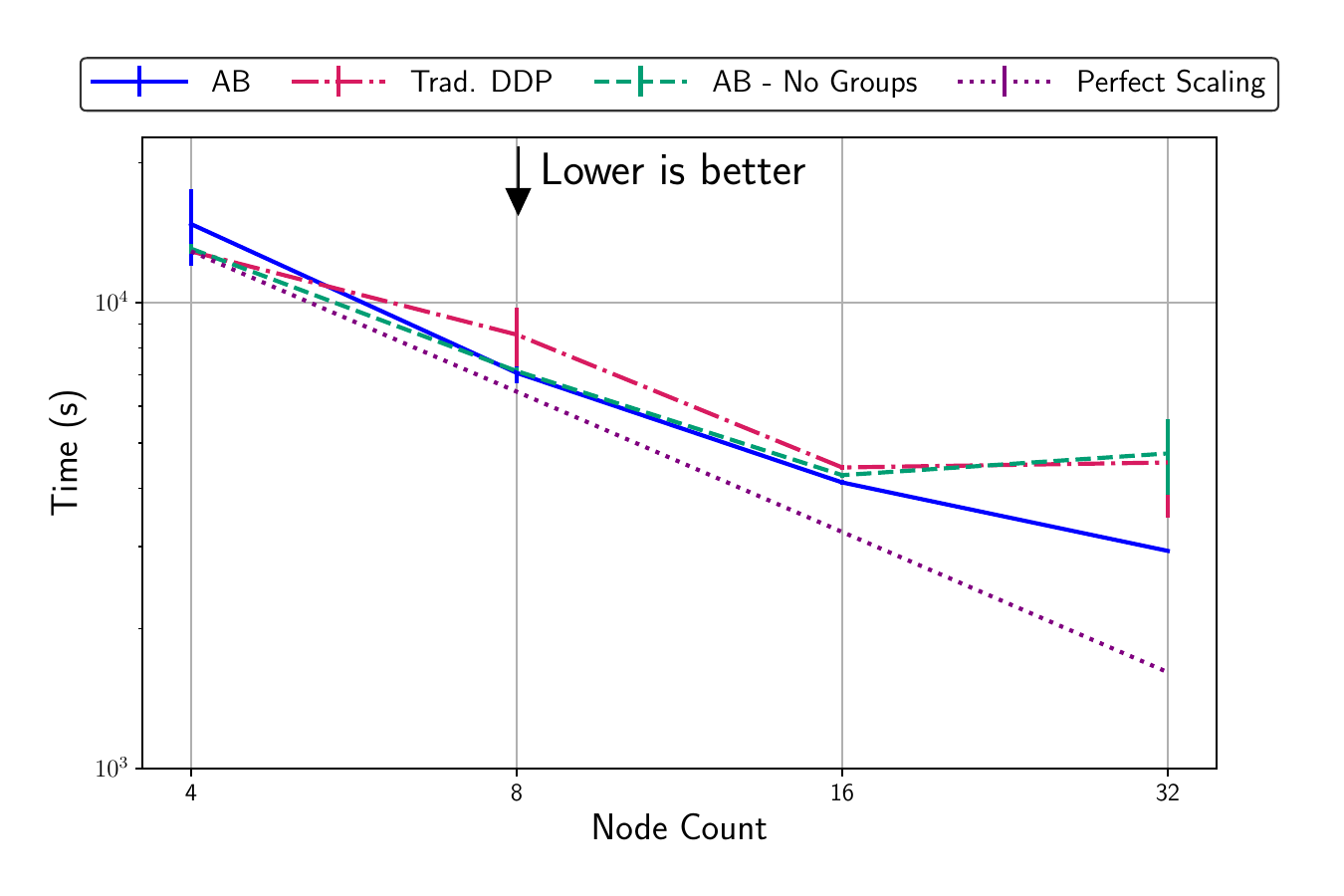}%
        \caption{Training wall-clock time.}%
        \label{fig:weak-time}%
    \end{subfigure}%
    \caption{Scaled interconnect traffic and job wall-clock time for the ViT B/16 trained on ImageNet-2012. Scaling is based on the average time required for the calculation and communication of gradients on 4 nodes.}
    \label{fig:weak-traffic-time}
\end{figure}

This experiment serves to show the performance of AB training as the time required to process a local batch shrinks, as is the case when the accelerators on a system are replaced but the network infrastructure remains the same.
It also provides insight into the ability of independent training groups to learn meaningful representations when the batch size used for independent training is much smaller than the global batch size. 
Without the independent groups, the accuracy of the trained network should be similar throughout scaling.
Results for these runs are shown in \Cref{fig:top1s-weak,fig:compress-weak,fig:weak-traffic-time}.

\section{Discussion}

Table \ref{tab:scaling} shows that our AB training method consistently achieves an average reduction of 70.31\% in scaled network traffic across the scaling tests.
The scaled network traffic is calculated as the average network traffic across the entire job scaled by the percentage of time required by the backward step during training (measured as 53.84\% on four nodes).
The traffic reduction highlights AB training's effectiveness in mitigating the DP training communication bottleneck, a key concern in distributed training, especially for large models in HPC environments.
These results underscore the significant bandwidth requirements for training even moderately sized neural networks, a challenge further exacerbated by larger models.

AB training maintains competitive accuracy compared to traditional DP training in 13 of the 18 scaling experiments (\Cref{fig:top1s-strong,fig:top1s-weak}), using the same hyperparameters and in a similar time frame.
Removing the independently trained groups noticeably reduces accuracy (\Cref{fig:vit-strong-top1}), while compression values remain similar (\Cref{fig:compress-strong}).
This demonstrates that AB training's performance benefits stem from both the low-rank representation and the independent subgroup training.

Our large-scale experiments reveal the interplay between communication efficiency and model accuracy.
In our ViT experiments with 32 nodes and a constant global batch size (\Cref{fig:weak-traffic-time}), we observe the aforementioned communication bottleneck, where the training time no longer decreases despite additional compute resources.
However, AB training continues to see a reduction in time to train due to its reduced communication demands, even with the additional overhead stemming from the computation of the SVD for all model weights.
ResNet-50's performance plateaus earlier due to its computation becoming the bottleneck, though AB training and the traditional DP baseline have similar accuracy and require roughly the same time to train.
AB training maintains or reduces the training time as compared to PyTorch's DDP for all scaling measurements (see \Cref{tab:scaling}).
This is most obvious in the measurements shown in \Cref{fig:weak-traffic-time}, .

Across all experiments, we observed favorable compression ratios for AB training(\Cref{fig:compress-weak,fig:compress-strong} and Table \ref{tab:scaling}).
For runs which equaled or exceeded the baseline network accuracy, the Vision Transformer's compression ratios range from 1.89 : 1 to 2.54 : 1.  
ResNet-50's compression was less impressive, although still respectable (ranging from 1.19 : 1 to 1.72 : 1 for outperforming runs).
This suggests a potential interaction between model architecture and the effectiveness of our method.

\Cref{tab:comparison} demonstrates AB training's effectiveness compared to existing low-rank and pruning methods. 
We showcase two standard benchmarks for these methods: ResNet-50 on ImageNet-2012 and VGG16 on CIFAR-10.
The latter highlights the potential of compression techniques due to VGG16's over-parameterization and CIFAR-10's relative simplicity \cite{chang2023icpprune}. 

AB training outperformed expectations on CIFAR10, achieving a 44.14 : 1 compression ratio (reducing the model to 2.27\% of its original size) with negligible accuracy loss. 
This surpasses the compression achieved by ICP~\cite{chang2023icpprune} (3.67\% of original size) and ABCPrune~\cite{lin2020abcpruning} (11.26\%).
For the ResNet-50 benchmark, AB training was the only one of the listed methods to outperform the baseline.

To compare the communication required for training the other model compression methods shown in \Cref{tab:comparison} with traditional DP, we estimate their maximum communication savings achievable.
As all of these methods iteratively reduce the network size, we assume that they all start with the full-rank representation and remove parameters during training.
In our experiments, we found that once the network began to compress the network, it removed much of the network quickly, then removed small amounts during the remaining training steps.
However, we do not know the rate at which parameters are removed.
Taking this into account, we assume that models train close to full rank for 25\% of training then close to their final model state for 75\% of training.
We describe the estimated communication reduction (ECR) as
\begin{equation} \label{eq:ecr}
    \text{ECR} = 100\% - F - Lc
\end{equation}
where $F$ and $L$ are the percentages of training spent near full-rank and the most-compressed network state respectively and $c$ is the final compression ratio expressed as a percentage.
For AB training, the percentage of training spent near the final compression state, $L$, is reduced by the percentage time spent in the independent group training phase.

\comparisontable 

AB training demonstrates the greatest communication reduction in the ResNet-50 benchmark with the ECR.
In practice, we measured an approximate network traffic reduction of 70\% for all use-cases, landing quite close to the ECR.
With the ECR, we outperform all models except OIALR and DLRT, which reduce the number of trainable parameters by training the $\Sigma$ matrix from a network's weight's SVD.
In the VGG16 scenario, ICP~\cite{chang2023icpprune} and AB training without using individual groups show slightly higher estimated communication reductions than AB training itself. 
This suggests that AB training offers superior communication efficiency until models reach extremely high compression levels.

Our results also reveal a complex interplay between low-rank representations, large batch sizes, and generalization.
We observe significant generalization improvement in all experiments, likely due to the regularization effect of low-rank representations and independent group training (see `No Groups' measurements in \Cref{fig:top1s-weak}).
However, at larger scales, accuracy degrades.
The decreasing compression ratios with increasing batch size in constant local batch size experiments (\Cref{fig:compress-strong}) suggest this degradation might be due to the loss or underutilization of important smaller singular values during model averaging.

Similarly, in constant global batch size experiments, increasing compression is coupled with decreasing accuracy (\Cref{fig:compress-weak,fig:top1s-weak}).
This is attributed to larger divergences in smaller singular values as local groups explore the loss landscape independently.
As this effect strengthens, the divergences between independent groups occur primarily in the smaller, yet important, singular values.
Averaging them together causes the divergent singular values to shrink relative to the other values.
Over the course of training, these values can become so small that they are removed, hindering AB training at extreme scales.


\scalingtable

Understanding the interplay between network bandwidth limitations and our method's potential benefits in real-world environments is crucial.
The observed accuracy degradation at scale suggests the need to research improved update mechanisms, potentially exploring non-average update rules, mixing matrices, or loss-based weighted averaging schemes.

\section{Conclusion}

Our experimental results demonstrate the significant potential of our AB training method to reduce interconnect traffic by utilizing low-rank representations and independently trained worker subgroups.
The consistent 70\% reduction in network traffic without an increase in training time nor a decrease in the model's accuracy has the potential to unlock new possibilities within distributed-memory computing environments. 
This could be particularly impactful for scientific research on machines without high-speed interconnects and for exploring even larger, more complex models.
Additionally, the pronounced regularization effects observed at smaller scales offer a promising direction for improving generalization and performance.

While our method successfully addresses communication challenges, its performance at extreme scales highlights the complex interplay between low-rank representations, large batch effects, and hyperparameter optimization.
Our findings emphasize the need for further research into tailored hyperparameter strategies and novel update mechanisms to fully harness the potential of this approach in massively scaled scenarios.

This work represents a significant step towards communication-efficient distributed training.
Our results offer valuable insights and highlight promising areas for future investigation. 
By addressing the limitations uncovered in our study, we believe the research community can pave the way for even more efficient and scalable training of large-scale neural networks, ultimately furthering scientific discovery across a variety of domains.

\section*{Acknowledgment}
This work was performed on the HoreKa supercomputer funded by the Ministry of Science, Research and the Arts Baden-W\"urttemberg and by the Federal Ministry of Education and Research.
This work is supported by the Helmholtz Association Initiative and Networking Fund under the Helmholtz AI platform grant and the HAICORE@KIT partition.

\bibliographystyle{IEEEtran}
\bibliography{references}

\end{document}